\documentclass[review]{elsarticle}

\usepackage{amssymb}
\usepackage{arabtex}
\usepackage{url}
\usepackage{utf8}
\usepackage[utf8x]{inputenc}
\usepackage[T1]{fontenc}
\usepackage{subfig}
\usepackage{algorithm2e}
\usepackage{longtable}
\novocalize
\begin{document}

\begin{frontmatter}

\title{A Hybrid Persian Sentiment Analysis Framework: Integrating Dependency Grammar Based Rules and Deep Neural Networks}

\author[label1]{Kia Dashtipour}
\address[label1]{Department of Computing Science and Mathematics, University of Stirling, Stirling, UK}

\author[label4]{Mandar Gogate}
\cortext[cor1]{Corresponding author: Mandar Gogate}
\ead{M.Gogate@napier.ac.uk}
\address[label4]{School of Computing, Edinburgh Napier University, Edinburgh, UK}

\author[label1]{Jingpeng Li}

\author[label6]{Fengling Jiang}
\address[label6]{Institute of Intelligent Machines, Chinese Academy of Sciences, Hefei, China}

\author[label6]{Bin Kong}
\author[label4]{Amir Hussain}

\begin{abstract}
Social media hold valuable, vast and unstructured information on public opinion that can be utilized to improve products and services.
The automatic analysis of such data, however, requires a deep understanding of natural language.
Current sentiment analysis approaches are mainly based on word co-occurrence frequencies, which are inadequate in most practical cases.
In this work, we propose a novel hybrid framework for concept-level sentiment analysis in Persian language, that integrates linguistic rules and deep learning to optimize polarity detection.
When a pattern is triggered, the framework allows sentiments to flow from words to concepts based on symbolic dependency relations.
When no pattern is triggered, the framework switches to its subsymbolic counterpart and leverages deep neural networks (DNN) to perform the classification.
The proposed framework outperforms state-of-the-art approaches (including support vector machine, and logistic regression) and DNN classifiers (long short-term memory, and Convolutional Neural Networks) with a margin of 10--15\% and 3--4\% respectively, using benchmark Persian product and hotel reviews corpora.
}
\end{abstract}

\begin{keyword}
Persian Sentiment Analysis \sep Low-Resource Natural Language Processing \sep Dependency-based Rules \sep Deep Learning
\end{keyword}

\end{frontmatter}

\section{Introduction}
\label{sec1}

In the era of digital media, e-commerce and social networks, websites allow users to share opinions and feedback about products and services.
Customers can make informed decisions by reading the experiences of other users \cite{yang2018sentiment, lauren2018generating}.
In addition, customer feedback can be used by the organizations to further improve the offered services.
For example, a traveller planning a holiday can search over the Internet to book a hotel, and the reviews can provide information related to the value for money, location, cleanliness and service. This can assist in making an informed decision and provide hotel management an opportunity to take into account the complaints made by past customers for providing a better service.
As another example, if someone wants to buy a laptop, the buyer may look at online reviews for understanding limitations such as laptops battery life, portability, build quality and usability etc., before purchasing.
Moreover, understanding the usability issues faced by laptop customers allow the manufacturer an opportunity to further ameliorate their laptops.
However, the quintillion bytes of data generated per day consisting of user feedback cannot be manually read and analyzed by an individual or an organization, for gauging public opinion ~\cite{poria2012merging,singh2013sentiment}.

Sentiment analysis is an automated process of computationally understanding and classifying subjective information from source materials such as reviews on e-commerce websites, and posts/comments on social media platforms.
A fundamental task in sentiment analysis is assigning polarity (positive or negative) to a given text.
However, an online review is generally a mixture of positive and negative comments about different aspects of products or services instead of expressing a biased positive or negative opinion.
For example, ``The restaurant serving staff were incredibly kind but the food was a bit cold when it came out''~\cite{mukherjee2012feature} expresses an positive sentiment towards the working staff and a negative sentiment towards the food serving.

Most of the current sentiment analysis approaches in Persian are based on word co-occurrence frequencies, that first extract frequency of the words containing polarity, and the overall polarity of source text is then determined.
However, these approaches fail to consider the word order and the dependency relation between words, that play an essential role in determining the overall polarity of the sentence.
For example, in ``Certainly old, but I`d love to see more stories in that universe'' though the reviewer is expressing a strong negative sentiment in the first part of sentence, the overall polarity is positive. Although, the words ``old'' and ``love'' have negative and positive polarities, the overall polarity not only depends on the word sentiment strength but also on the relative words position and dependency structure of the sentence~\cite{leung2009sentiment,poria2016convolutional}.

Moreover, current approaches of sentiment analysis fail to understand the real noisy text data consisting of sarcasm, idioms, informal words, phrases and spelling mistakes~\cite{cambria2018senticnet,dashtipour2016multilingual,dashtipour2017adaptation,hussien2018comparison}. For example, \RL{by tfAwt nystm fq.t dygr ksy brAym mhm nyst}, a sarcastic sentence, cannot be accurately classified by the current approaches \cite{nodoushan2008conversational}.
In addition, the scarce availability of natural language processing tools and resources such as lexicon, labelled corpus, parts-of-speech (POS) tagger, etc.
are a major bottleneck in the reliable implementation of sentiment analysis methods in Persian, a major language with 54 million speakers ~\cite{ghayoomi2012word,dashtipour2017persian,dashtipour2017comparative}.

To address the aforementioned limitations, we propose a novel framework that integrates Persian linguistic grammar rules and deep learning for analyzing source text. This is shown to improve the overall performance and robustness of polarity detection for real noisy data.

Dependency grammar rules are based on linguistic patterns that allow sentiment to go from words to concepts based on dependency relations.
As a result, dependency-based rules take into account hierarchical relations between keywords, the word order, and individual word polarities to accurately determine the underlying polarity.

\begin{figure}[!t]
\centering
\includegraphics[width=\textwidth]{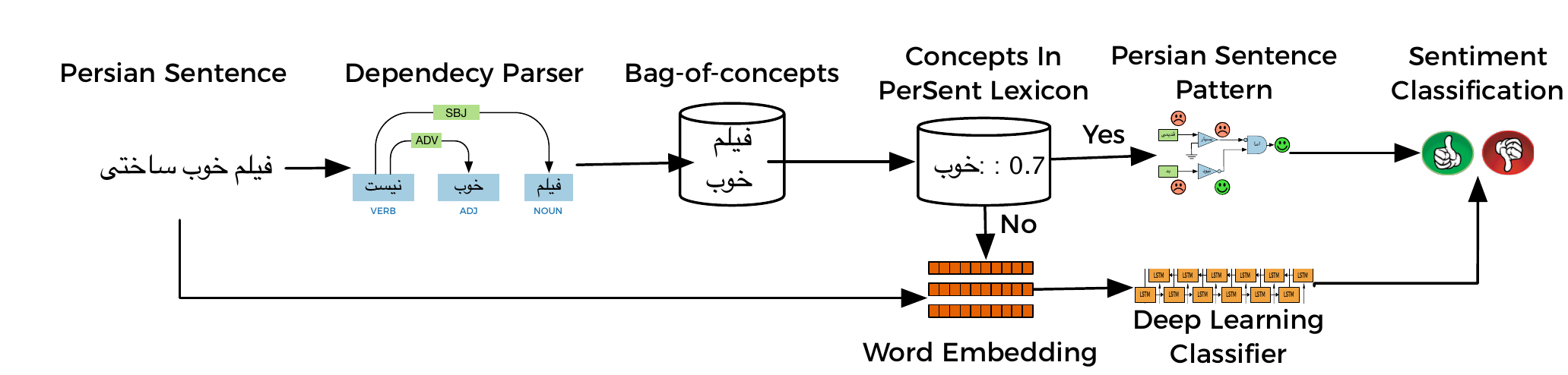}
\caption {Proposed Framework overview: Integrating Dependency-Based Rules and Deep Neural Networks for Persian Sentiment Analysis}
\label{fig:one12}
\end{figure}

We perform an extensive and comprehensive set of experiments using benchmark Persian product and hotel reviews corpora and compare the performance of our proposed dependency-grammar rule-based approach with state-of-the-art Support Vector Machine (SVM), Logistic Regression (LR), and DNN architectures including long short-term memory (LSTM) and Convolutional Neural Networks (CNN).
Comparative simulation results reveal that dependency-based rules outperform state-of-the-art SVM, LR, and fastText classifiers by a margin of 10-15\% and perform comparably with DNN classifiers.
However, the dependency-based rules cannot classify 10-15\% of the dataset due to non-availability of word polarity in the small sized Persian lexicon.
Therefore, we propose a hybrid framework, shown in Fig.~\ref{fig:one12}, that integrates our proposed dependency rule-based classifier and different DNN architectures, such as CNN and LSTM.

In summary, the paper reports four major contributions outlined below:
\begin{enumerate}
\item Novel dependency-based rules for Persian sentiment analysis. These address the limitation of word co-occurrence frequency based approaches by exploiting the hierarchical relations between keywords, the word order and, individual word polarities to accurately determine the underlying sentiment. To the best of our knowledge, dependency rule based classifier is first of its kind that assigns polarity to Persian sentences without any supervised training algorithm.
\item A critical analysis of our proposed dependency-based rules with conventional Logistic Regression, Support Vector Machine as well as advanced DNN models, on two benchmark product and hotel reviews corpora, demonstrates the superior performance of our proposed model compared to conventional approaches and advanced DNN models.
\item An ablation study of our proposed dependency-based rules reveals the importance of individual rules in the context of the complete framework.
\item Addressed limitations of unclassified sentences with our proposed dependency-based rules approaches by proposing a hybrid framework that integrates Persian dependency-based rules and DNN models, including CNN and LSTM to further improve the performance.
\end{enumerate}

The rest of the paper is organized as follows: Section 2 presents related work on state-of-the-art sentiment analysis approaches for English, Persian and other languages.
Section 3 presents our proposed dependency-based rules for Persian sentiment analysis.
Section 4 describes our proposed hybrid framework to address the limitations of dependency-rule driven models, and architectural details for DNN models, including CNN and LSTM, used in our hybrid framework.
Section 5 discusses comparative experimental results and ablation studies.
Finally, section 6 concludes this work with limitations of our current approach and outlines future research directions.

\section{Related Work}\label{sec:related-work}

In the literature, extensive research has been carried out to build different sentiment analysis models for English, Persian and other languages approaches.

\subsection{English Sentiment Analysis Approaches}\label{subsec:english-sentiment-analysis-approaches}

Sentiment analysis techniques can be broadly categorized into symbolic and sub-symbolic approaches: the former include the use of lexicons~\cite{banlex}, ontologies~\cite{draont}, and semantic networks~\cite{camnt5} to encode the polarity associated with words and multi-word expressions;
the latter consist of supervised~\cite{onesta}, semi-supervised~\cite{hussem} and unsupervised~\cite{liilea} machine learning techniques, that perform sentiment classification based on word co-occurrence frequencies.
Among these, the most popular algorithms are based on deep neural networks~\cite{yourec} and generative adversarial networks~\cite{liigen}.
There are also a number of hybrid frameworks that leverage both symbolic and sub-symbolic approaches.
Specifically, Jia et al.~\cite{jia2009effect} proposed lexical rules to handle the problem of determining the sentiment when negation is present in the sentences.
Similarly, Taboada et al.~\cite{taboada2011lexicon} use dictionaries of words annotated with semantic relations that incorporates phenomena such as intensification, negation or adversatives to determine the polarity of source text.
Thelwall et al.~\cite{thelwall2012sentiment} used a lexicon and rules to detect polarity in short informal English texts to handle negation.
In addition, Poria et al.~\cite{poria2014sentic} proposed a set of linguistic rules to propagate polarity through dependency trees.
Finally, Agarwal et al.~\cite{agarwal2015concept} presented a novel feature extraction technique based on the dependency-based semantic parsing and ConceptNet ontology. The extracted features are fed to a SVM classifier to categorize a sentence into positive or negative.

\subsection{Multilingual Sentiment Analysis Approaches}\label{subsec:other-languages}

Boiy et al. \cite{boiy2009machine} proposed an approach to detect polarity in English, Dutch and French texts found on social media using n-gram features. Experimental results showed that unigram features outperformed bigram features.
Balahur et al. \cite{balahur2014comparative} developed a machine translation for languages such as French, German and Spanish with scarce resources for the sentiment analysis task. Empirical results showed that their proposed approach depends on the accuracy of the translation engine and the system was unable to detect sentiment when a mixture of languages is fed as input. The main limitation with translation based sentiment analysis approaches for scarce resource languages is that, current state-of-the-art machine translation systems fail to provide contextual translation in subjective sentences. In addition, the output of the translation system may not reflect the actual writing styles that exist in real life.

\subsection{Persian Sentiment Analysis Approaches}\label{subsec:persian-sentiment-analysis}

Basiri et al.~\cite{basiri2017sentence} addressed the problem of lack of resources in Persian by developing a lexicon and an automatically labelled sentence-level corpus.
A Naive Bayes classifier was trained on the collected corpus to determine the polarity in short sentences.
In addition, Ebrahimi et al.~\cite{ebrahimi2017supervised} proposed a method to detect polarity in Persian online reviews based on adjectives extracted from the sentence and translated SentiWordNet lexicon.
However, the proposed method is limited to adjectives and it does not exploit nouns, adverbs and verbs that provide extra information on the underlying sentiment.
Moreover, Razavi et al.~\cite{razavi2017word} detect polarity based on extracted nouns and adjectives and a Persian lexicon. However, the proposed approach do not exploit the word order and hierarchical semantic dependency.

Aleahmad et al.~\cite{aleahmad2007n} proposed a method to detect polarity based on n-gram features.
Experimental findings showed that fourgram outperforms unigram, bigram and trigram features.
On the other hand, Dashtipour et al. \cite{dashtipour2018exploiting} proposed deep autoencoder based feature extraction for sentiment analysis.
The framework outperformed state-of-the-art CNN and multilayer perceptron (MLP) for detecting polarity in Persian text.

Most of the aforementioned studies exploit word co-occurrence frequencies and a lexicon to determine polarity of source text. These generally fail to exploit hierarchical semantic relations, and word order.
In addition, current studies that use lexical rules to detect negation in various languages cannot be directly applied for Persian.
Moreover, deep neural networks have shown state-of-the-art performance using a large supervised corpus for various natural language processing tasks including sentiment analysis.
However, for scarce resource languages, we need innovative methods that jointly exploit deep learning models (sub-symbolic) and dependency based rules (symbolic) approaches to go beyond current state-of-the-art performance.
Therefore, we propose a novel hybrid framework for Persian sentiment analysis, that integrates dependency-based rules, and deep neural networks to improve the overall performance and robustness of polarity detection in real noisy data.

\section{Methodology}\label{sec:methodology}

\begin{figure}[!t]
\centering
\includegraphics[width=0.3\textwidth]{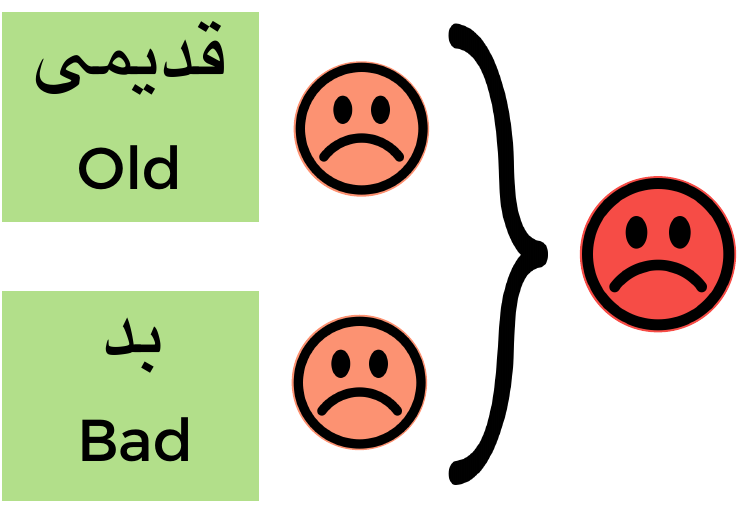}
\caption {Traditional approaches extract words ``old'' and ``bad'' from a sentence and assign negative polarity into the sentence}
\label{fig:five-two}
\end{figure}

\begin{figure}[!t]
\centering
\includegraphics[width=\textwidth]{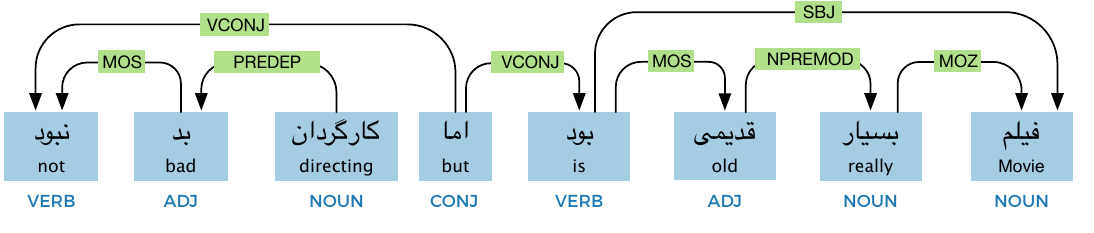}
\caption {Dependency tree of a sentence (SBJ - Subject, MOZ - Extra Dependent, NPREMOD - Pre-Modifier of Noun, MOS - Mosnad, VCONJ - Conjunction of Verb, PREDEP - Pre-Dependent, ADJ - Adjective)}
\label{fig:five-three}
\end{figure}

\begin{figure}[!t]
\centering
\includegraphics[width=0.5\textwidth]{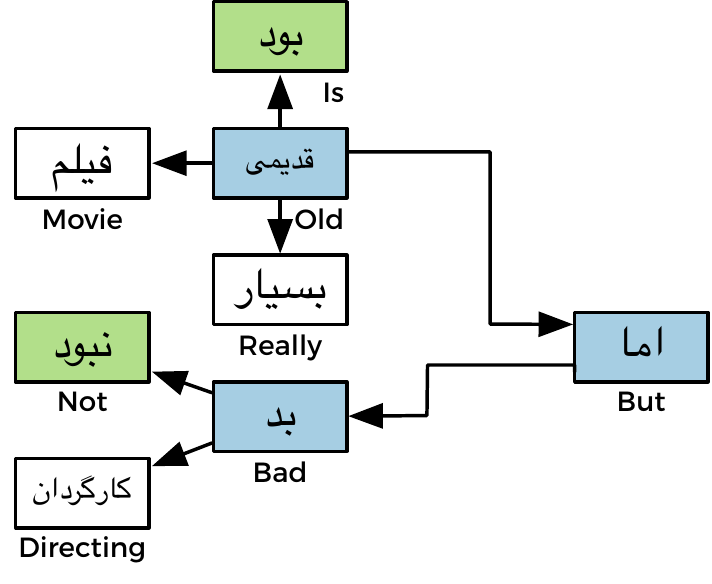}
\caption {The dependency tree of a sentence resembles an electronic circuit}
\label{fig:five-four}
\end{figure}

\begin{figure}[!t]
\centering
\includegraphics[width=0.5\textwidth]{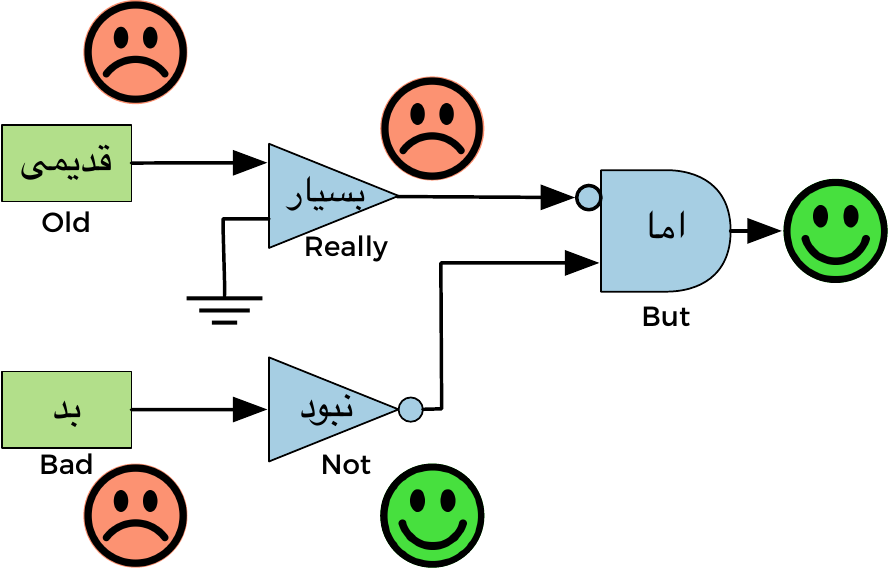}
\caption {The final sentiment data flow of the signal in the circuit}
\label{fig:five-five}
\end{figure}

This section describes our proposed dependency-based rules for Persian sentiment analysis.
The proposed rules exploit hierarchical dependency relations to more accurately determine the underlying sentiment, compared to traditional word co-occurrence frequency based approaches.
For example, when using frequency of positive and negative words to classify a sentence like ``The movie is very old but directing is not bad'' (\RL{fylm bsyAr qdymy bwd AmA kArgrdAny bd nbwd}), the sentence will be classified as negative since the sentence consists of two negative words (``old'' \RL{qadymy} and ``bad'' \RL{bd}) , as shown in Fig \ref{fig:five-two}. However, the actual polarity of the sentence is positive due to the dependent words (i.e. ``not'' and ``but"), which change the overall polarity of the sentence to positive.
On the other hand, our proposed dependency-based rules take into account the syntactic relation between words, as shown in Fig \ref{fig:five-three} and \ref{fig:five-four}.
The exploitation of the dependent words in dependency-based rules establish a logical flow of sentiment, as shown in Fig \ref{fig:five-five}, to determine the overall polarity.
Specifically, the word ``old'' \RL{qdymy} following ``very'' \RL{bsyAr} does not change the overall polarity, but the negative word ``bad'' \RL{bd} following negation changes the overall polarity of the sentence into positive.
Finally, the use of word ``but'' \RL{AmA}  changes the polarity of the second component of the sentence.

\subsection{Persian Dependency-based Rules}\label{subsec:other-rules}

This section presents novel dependency-based rules for sentiment analysis that are adopted from~\cite{windfuhr2011persian} and~\cite{elwell1972elementary}.

\subsubsection{Polarity Inversion}

\textbf{Trigger}: When a sentence consists of at least one negation word such as \RL{nbwd}, \RL{nyst}

\textbf{Action}: The negation in a sentence can change the polarity of the sentence.
For example, if negation is used with a positive token, the overall polarity of the concept is negative and if negation is used with a negative token, the overall polarity of the concept is positive.
For example, ``I do not like the Samsung mobile'' (\RL{mn mwbAyl sAmswng dwst ndArm}), the overall polarity of this sentence is negative.

\subsubsection{Complement Clause}

\textbf{Trigger}: The complement clause is introduced by ``that'' \RL{kh}  in the sentence.
Usually, the complement clause is connected by noun, verb and adjective in the sentence. %

\textbf{Action}: The sentence is split into two parts based on the complement clause and the first part of the sentence is considered to identify the overall polarity of the sentence.
For example, ``I am happy that you did not buy old mobile'' (\RL{mn xw^s.hAlm keh Ayn} \RL{mwbAyl qdymy nxrydy}), the sentiment expressed before the word ``that'' \<keh> is positive and the overall polarity of the sentence is positive.

\subsubsection{Adversative}

\textbf{Trigger}: The adversative can be a word, phrase or clause implying opposition or contrast in the sentence.
An adversative word like ``but'' \RL{AmA}, ``although'' \RL{Agr ^ceh}, or ``however'' \RL{bA Aynkeh} is used to connect two elements of opposite polarities.

\textbf{Action}: The sentences are split into two parts based on the conjunction such as ``but'' and the polarity of the second part is considered. For example, in the sentence ``The iPhone is really good but it`s very expensive'' (\RL{mwbAyl Apl xyly xwb Ast wly xyly grAn Ast}), the first part before the word ``but'' is positive and the second part is negative. Hence, the overall polarity of the sentence is negative. Moreover, for ``The Apple mobile is really expensive but it`s very good'' (\RL{mwbAyl Apl xyly grAn Ast AmA xyly xwb Ast}), the first part of the sentence is negative and the second part is positive.
Therefore, the overall polarity of the sentence is positive.

\subsubsection{Adverbial Clause}

\textbf{Trigger}: When a sentence contains an adverbial clause.
The adverb clause is a group of words which count as an adverb in the sentence.
The adverb clause must contain a subject and verb to trigger the rule.

\textbf{Action}: The role of ``whereas'' in a sentence is like the word ``but''. If a sentence contains ``whereas'', the sentence is split into two parts by identifying the subject in the sentence, since the overall sentiment is identified by the second part of the sentence.
For example, ``In the product description they said the mobile has good lens whereas the lens is not good'' (\RL{dr tw.dy.haat m.h.s^ul gfteh bwd keh mwbAyl lnz xwby dr .swrtykeh keh lnz^s lrz^s dArd}).
The polarity of the first part is positive and the polarity of the second part (after the whereas) is negative.
Hence, the overall polarity of the sentence is negative.

\subsubsection{Adjective Clause}

\textbf{Trigger}: When a sentence contains an adjective clause.
The rule is triggered when the sentence consists of pronouns such as ``which'' \RL{keh, Ayn}.

\textbf{Action}: The role of ``which'' \RL{keh, Ayn} is similar to the rule of ``but'' in the sentence.
In this instance, the sentence is split into two parts and the polarity of the second part is considered.
For example, ``Read about things which are beautiful and good'' (\RL{xwndn dr mwrd msAyly kh Ayn hmh zybA w xwb hstnd})

\subsubsection{Joint Noun and Adjective}

\textbf{Trigger}: When a sentence contains joint nouns and adjectives.

\textbf{Action}: When there is a relation between noun and adjective, both of these words are extracted from the sentence and the polarity of adjectives are considered.
For example, ``The mobile is bad'' (\RL{Ayn mwbAyl bd bwd}).
There is a subject relation between ``mobile'' \RL{mwbAyl} and ``bad'' \RL{bd}.

\subsubsection{Preposition}

\textbf{Trigger}: When a sentence contains a preposition.

\textbf{Action}: The preposition ``against'' \RL{mxaalf} is generally used in negative sentences.
However, it can also be used in the positive sentences.
Usually when an activity contains a negative sentiment and it follows a negative preposition modifier, the overall polarity of the sentence is changed to positive.
For example, ``I am against this request'' (\RL{mn mxaalf Ayn drxwAst hstm}).
On the other hand, if an activity is positive and it follows a negative preposition modifier, the overall polarity of the sentence is changed to negative.
For example, ``Hitler raised a great army and made war against other countries'' (\RL{hytlr Art^s bzrgy sAxt w jngy `lyeh k^swrhAy mxAlf rA ^srw` krd}).

\subsubsection{Additional rule}

\textbf{Trigger}: When a sentence contains the word ``This'' \RL{Ayn} in the middle of the sentence.

\textbf{Action}: The sentence is split into two parts based on the appearance of word (\RL{Ayn}) in the sentence, and the first part is considered.
For example, ``I had LG mobile, it was very bad, anyway, this mobile was for my brother'' (\RL{mn qblA mwbAyl Aljy dA^stm xyly bd bwd bh hr .hAl Ayn mwbAyl mAl brAdrm bwd}), the sentiment expressed after the word ``this'' \RL{Ayn} does not contain any sentiment. However, the sentence contains negative words ``I am not sure'' which changes the polarity of the sentence into negative. The negative polarity is assigned to the sentence because in the first part, the reviewer is not sure about the quality of the mobile.

Table \ref{tab:five-two} summarized the proposed dependency grammar based rules for Persian.
\begin{longtable}[!t]{| p{.15\textwidth} | p{.80\textwidth} |}
\caption{Overview of Dependency-Based Rules}
\\\hline
\centering
\label{tab:five-two}

\textbf{Rules} & \textbf{Behaviour} \\
\hline
Polarity Inversion & \vspace{-1mm} In this rule, if negation is used with a positive token, the polarity is negative and if negation is used with a negative token, the polarity is positive \\
\hline
Complement Clause & \vspace{-1mm} In this rule, the sentiment expressed before the word ``that'' (\RL{keh}) considered \\
\hline
Adverbial Clause & \vspace{-1mm} In this rule, the sentence is split into two parts and the polarity of the second part is used as the overall polarity of the sentence. \\ \hline
Adjective Clause & \vspace{-1mm} In this rule, if the word ``which'' (\RL{keh Ayn}) is in the sentence, the sentence is split into two parts and the polarity of the second part is considered \\ \hline
Joint Noun and Adjective & \vspace{-1mm} In this rule, when there is a relation between noun and adjective, both noun and adjective are extracted. The lexicon is used to assign polarity to extracted words. \\
\hline
Adversative & \vspace{-1mm} In this rule, the word if a word like ``but'' (\RL{AmA}), ``although'' (\RL{Agr}), or ``however'' (\RL{bA Aynkeh}) is used in the sentence.
The sentences are split into two parts and the polarity of the second part is considered \\
\hline
Preposition & \vspace{-1mm} In this rule, if the word ``against'' (\RL{mxaalf}) is in the sentence.
It will change the polarity of the sentence into negative \\
\hline
Additional rule & \vspace{-1mm} In this rule, when the sentence contains the word ``This'' (\RL{Ayn}), the sentence is split into two parts and the first part of the sentence is considered the polarity of the sentence. \\
\hline
Preposition Sub-rule &  \vspace{-1mm} In this sub-rule, if positive prepositions are used before any adjective, they can change the polarity of the sentence into positive and if negative prepositions are used before any adjective, they can change the polarity of the sentence into negative. \\
\hline
Emoji Sub-rule &  \vspace{-1mm} In this sub-rule, if the positive emoji is appearing in the sentence, the polarity of the sentence is positive and if negative emoji is appearing in the sentence the polarity of the  sentence is negative. \\
\hline
\end{longtable}

\subsection{Preposition Sub-rule}\label{subsec:sub-rules}

The Persian sentences contain different prepositions that consist of polarity. If positive prepositions words ``with'' \RL{bA}, ``happy'' \RL{xw^s} and ``enjoy'' \RL{.sfAy}  are used before any adjective, they can change the polarity of the concept to positive. For example, \RL{mn xw^s.hAl hstm} ``I am happy today", because the word \RL{xw^s} is appearing in the sentence, the polarity of the sentence is positive. However, when some preposition words such as ``without'' \RL{by} ``anti'' \RL{.dd}, ``Not'', \RL{nA}, ``poison'' \RL{zhr} and ``No'' \RL{lA} appear in the sentence, the polarity of the sentence is changed into negative. For example, in ``Your answer is not wrong'' (\RL{jwAb ^smA nAdrst Ast}) the polarity is negative, since the word ``not'' appears in the sentence. The word \RL{nA} is not detected by the negation rule as \RL{nA} is not a negation word and the rule cannot be triggered.

\subsection{Emoji Sub-rule}\label{subsec:-emojisub-rules}

Emojis were introduced as expressive components in the sentence. Emojis generally reinforce the polarity expressed in the sentence, except in sarcastic sentences. 
Emojis can be divided into positive  :\--),:), :\--], :], :D or negative :(, :\--((, :`(. Online reviews consist of different emojis. For example, :)\RL{mn gw^sy tAzh xrydm, gw^sy xwbyh tw xryd^s ^sk ndA^sth bA^syd }, ``I recently bought the mobile, do not hesitate to buy this mobile :)``. Therefore, If a positive emoji is appearing in a sentence, the polarity of the sentence is positive and if a negative emoji is appearing in a sentence, the polarity of the sentence is negative.

    \begin{algorithm}[!t]
        \SetAlgoLined
        \KwResult{The polarity of the sentence}
        Tokenize and normalize sentences using hazm parser\;
         \For{word in sentence}{
         \eIf{word in lexicon}{
            assign opinion strengths to each extracted word\;
         }{
            assign zero polarity\;
         }
         }
        apply dependency-based rules\;
        \eIf{polarity assigned by dependency-based rules}{
        return polarity\;
        }{
        apply DNN classifiers\;
        return polarity\;
        }
        \caption{Proposed hybrid framework}
        \label{algor:hybrid}
    \end{algorithm}

\section{Hybrid Framework}\label{sec:classification}

In this section, we discuss our proposed hybrid approach that integrates dependency-based rules with DNN classifiers including CNN and LSTM.

\subsection{Framework Overview}

Our proposed hybrid approach combines deep learning and dependency-based rules to address the problem of unclassified sentences in the aforementioned rule-based approach.
The algorithm \ref{algor:hybrid} depicts an overview of our proposed hybrid framework.
First, sentences are preprocessed and the PerSent lexicon is used to assign polarity in extracted words.
The word polarities and the sentence dependency tree is fed to a dependency-based rules classifier.
If the rule-based classifier is unable to classify sentences either due to unavailability of subjective keywords in the lexicon, or if no rule was triggered, then the concatenated fastText embedding of the sentence is fed into a DNN classifier to determine the polarity of the sentence.

\subsection{Deep Neural Networks {DNN} architecture}

This section presents architectural details of DNN models used in our proposed hybrid framework, as well as standalone classifiers used for comparative performance evaluation. 
In recent years, DNN classifiers have achieved state-of-the-art performance as compared with other techniques due to their inherent ability to represent data at different levels of abstraction.
The primary advantage of using deep learning classifiers such as CNN and LSTM is that they do not require any manual feature engineering~\cite{dos2014deep}.
However, word representations learned using a relatively small supervised corpus perform poorly as compared to unsupervisely trained fastText word embeddings.
Therefore, we feed the concatenation of fastText embeddings, of size $(n \times k, 1)$ where n is the maximum number of words in a sentence and k is the embedding dimension), to the DNN architectures.

\begin{figure}[!t]
\centering
\includegraphics[width=\textwidth]{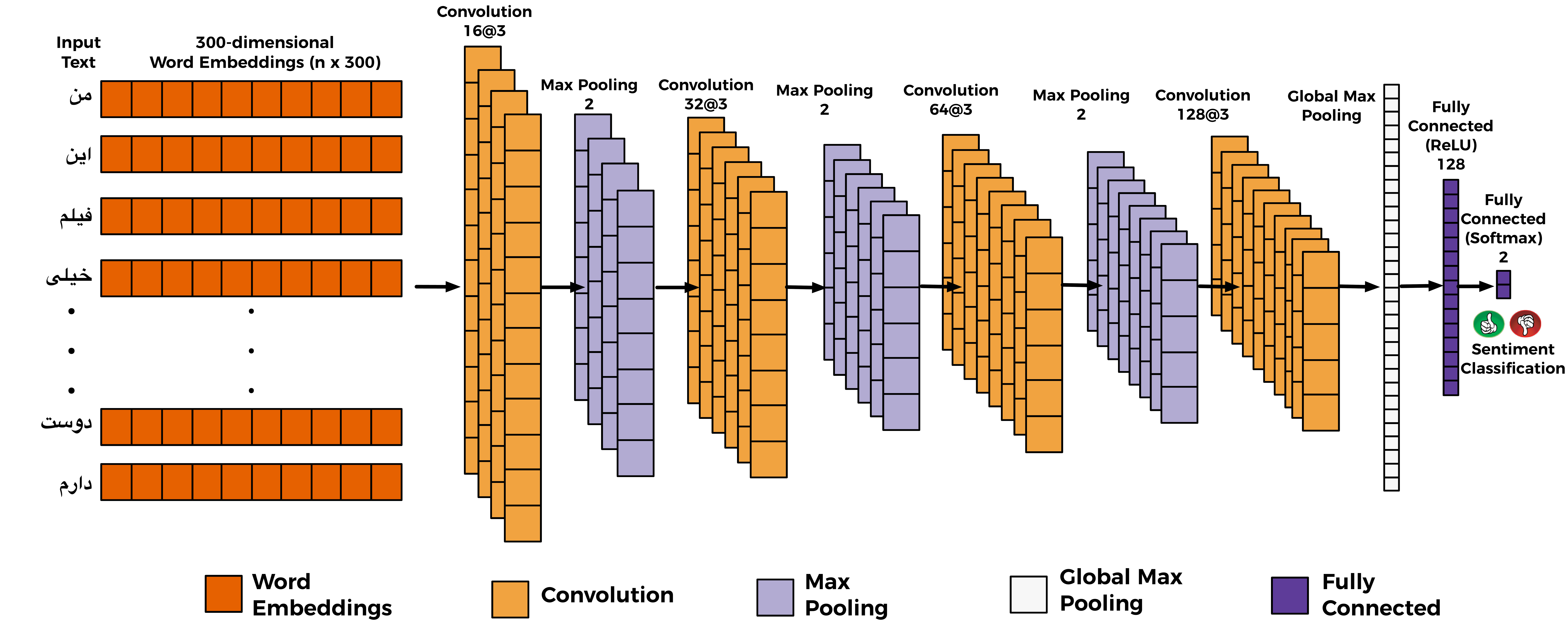}
\caption {Convolutional Neural Network}
\label{fig:five-six}
\end{figure}

\subsection{Convolutional Neural Network (CNN)}\label{subsec:convolutional-neural-networkcnn}
Our developed CNN architecture, is similar to the one used by Poria et al. ~\cite{poria2016fusing} and Gogate et al.~\cite{gogate2017novel,gogate2017deep}, consisting of input, hidden and output layers.
The hidden layers consist of convolutional, max pooling, and fully connected layers.
In our experiments, the best results are obtained with a 9-layered CNN architecture as illustrated in Fig \ref{fig:five-six} and Table \ref{cnn:architect12}

\begin{figure}[!t]
\centering
\includegraphics[width=\textwidth]{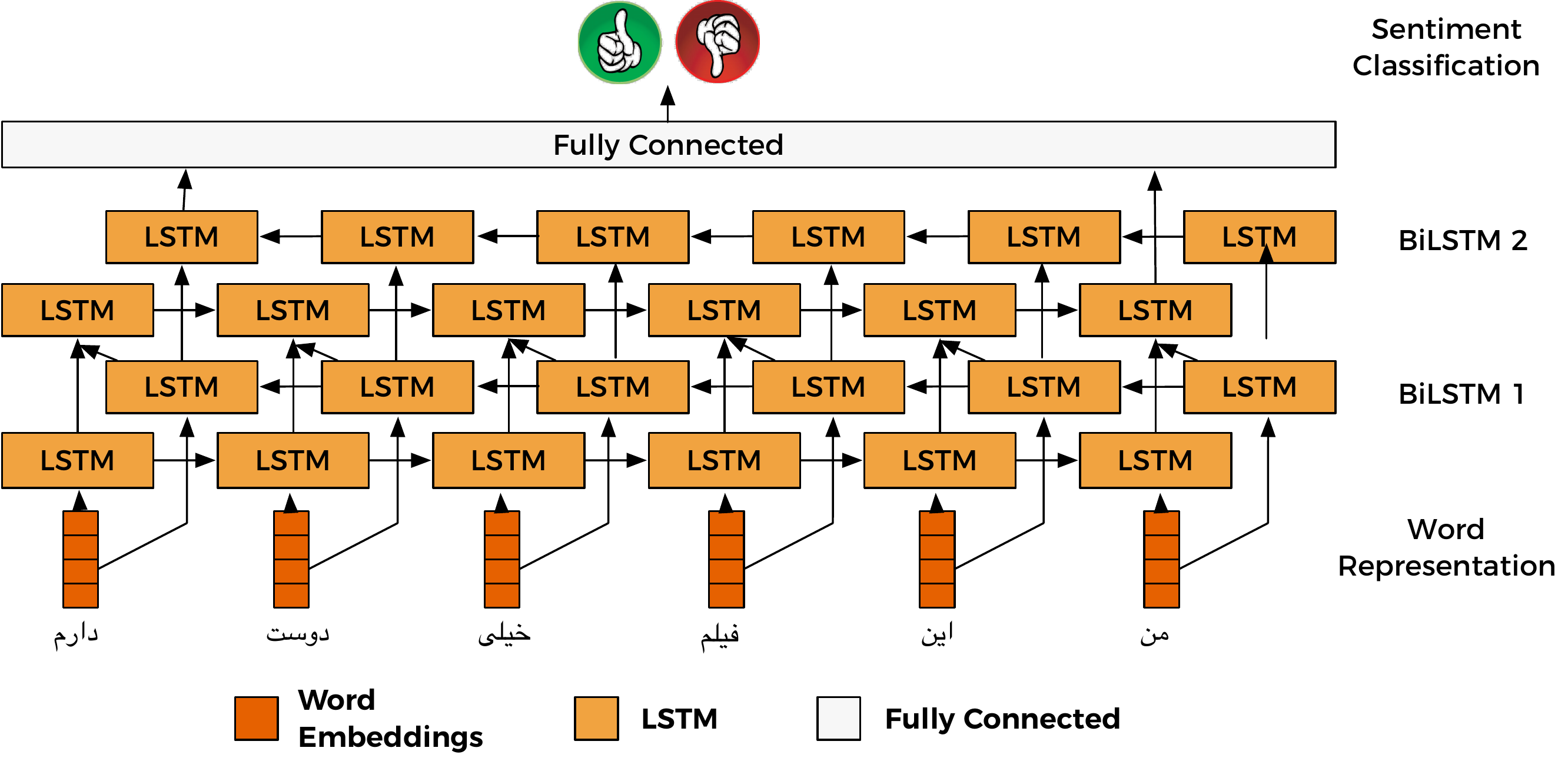}
\caption {Long Term Short Memory}
\label{fig:five-seven}
\end{figure}

\subsection{Long Short Term Memory (LSTM)}\label{subsec:long-short-term-memorylstm}
Our developed LSTM architecture, is similar to the one used by Wang et al.~\cite{wang2016attention}, consisting of an input layer, two stacked LSTM layers and one output fully connected layer.
Specifically, the LSTM part consists of two stacked bidirectional LSTM layers with 128 and 64 cells, followed by a dropout layer with 0.2 probability, and one dense layer with two neurons and softmax activation.
Our developed LSTM is illustrated in Fig \ref{fig:five-seven}.

\begin{table}[!t]
\centering
\caption{CNN Architecture (Conv - convolutional layer, MaxPool - Maxpooling layer, GlobalMaxPool - Global Max Pooling layer, Fc - Fully connected layer, ReLU - Rectified Linear Unit Activation}
\label{cnn:architect12}
\resizebox{\textwidth}{!}{%
\begin{tabular}{|l|l|l|l|l|l|l|l|l|l|l|}
\hline
Layer & 1 & 2 & 3 & 4 & 5 & 6 & 7 & 8 & 9 & 10 \\ \hline
Type & Conv & Max & Conv & Max & Conv & Max & Conv & Global & Fc & Fc \\ \cline{1-2} \cline{4-4} \cline{6-8} \cline{10-11} 
Filters & 16 & Pool & 32 & Pool & 64 &  & 128 & Max &  &  \\ \cline{1-8} \cline{10-11} 
Kernal Size & 3 & 2 & 3 & 2 & 3 & 2 & 3 & Pool &  &  \\ \hline
Neurons &  &  &  &  &  &  &  &  & 128 & 2 \\ \hline
Activation & ReLU &  & ReLU &  & ReLU &  & ReLU &  & ReLU & SoftMax \\ \hline
\end{tabular}%
}
\end{table}

\section{Performance Evaluation}\label{sec:performance-evaluation}

In this section we discuss the datasets, data preprocessing, experimental results and ablation studies.

\subsection{Dataset}

In this experiment, two benchmark Persian product and hotel reviews corpora are used to evaluate the performance.

\textbf{Product reviews dataset} \cite{basiri2018words}: The dataset consists of 1500 positive reviews and 1500 negative reviews collected from the product review website
(www.digikala.com).
There are two types of labels in the dataset (i.e. positive or negative) which are manually annotated.

\textbf{Hotel reviews dataset} \cite{alimardani2015opinion}: The dataset consists of 1800 positive and 1800 negative hotel reviews collected from hotel booking website (http://www.hellokish.com).
The hotel reviews corpus is used to compare how our approach performs in a new domain compared to state-of-the-art approaches, including multilingual methods.

\subsection{Data Preprocessing}

The corpus is preprocessed using tokenization and normalization techniques. The process of converting sentences into single words or tokens is called tokenization. For example, ``The mobile is great'' is broken down to ``The", ``mobile", ``is'' and ``great"~\cite{sumathy2013text}.
Normalization is used to convert the tokenized words with various suffixes into their respective normal  form~\cite{reynolds1997comparison}. For example, \RL{mrsy} ``Thanks'' is written as \RL{mr30}, or \RL{`Aly} ``great'' is written as \RL{`Alyyyyy}.
The numbers and punctuation were removed from the sentences.
In addition,  PerSent lexicon \cite{dashtipour2016persent} is used to assign polarity to the tokenized words. Zero polarity is assigned to the words that are not present in the lexicon.
Finally,  the Hazm Python package is used to identify the dependency tree for a sentence, the overall accuracy for hazm parser is 90.86\%~\cite{khallash2013empirical}.
The dependency tree and assigned polarities are fed to our proposed dependency-based rules classifier. 

\subsection{Experimental Setup}

Our proposed dependency-based rules do not require any training. However, to facilitate a fair comparison between dependency-based rules and other approaches we split the data into train (60\%), validation (10\%) and test set (30\%), and evaluated the rule-based approach on the test set.
For the rule-based approach alone, positive polarity is chosen when dependency-based rules are unable to classify the sentence mainly due to the unavailability of word polarity in the small sized Persian lexicon. In addition, SVM, Logistic Regression and fastText classifiers were used as a baseline to compare the performance of our proposed approach.
The DNN architectures were trained and validated using TensorFlow library and NVIDIA Titan X GPU. The models were trained for 200 epochs using back propagation, with the Adam optimizer minimizing the categorical cross entropy loss function. Unclassified sentences from the rule-based approach were converted into 300 dimensional pretrained fastText word embedding and fed into deep learning classifiers as a part of the hybrid framework.

\subsection{Experimental Results}

\begin{table}[!t]
\centering
\caption{Summary of the results for Product Reviews}
\label{tab:five-three}
\setlength{\tabcolsep}{1pt}
\begin{tabular}{|l|c|c|c|c|}
\hline
\textbf{Classifier} & \textbf{Precision} & \textbf{Recall} & \textbf{F-measure} & \textbf{Accuracy} \\ \hline
Kim et al.\cite{kim2014convolutional} & 0.60 & 0.63 & 0.60 & 62.5 \\ \hline
Dehkharghani et al.\cite{dehkharghani2012adaptation} & 0.60 & 0.93 & 0.72 & 63.62 \\ \hline
fastText Classifier\cite{joulin2017bag} & 0.67 & 0.67 & 0.67 & 70.01 \\ \hline
SVM & 0.75 & 0.75 & 0.75 & 74.8 \\ \hline
Logistic Regression & 0.75 & 0.75 & 0.75 & 75.2 \\ \hline
Dependency-Based Rules & 0.84 & 0.81 & 0.80 & 80.70 \\ \hline
CNN & 0.89 & 0.71 & 0.78 & 78.07 \\ \hline
LSTM & 0.90 & 0.74 & 0.81 & 79.77 \\ \hline
\begin{tabular}[c]{@{}l@{}}
Hybrid 1: CNN \\+ Dependency-Based Rules
\end{tabular} & 0.75 & 0.98 & 0.84 & 81.14 \\ \hline
\begin{tabular}[c]{@{}l@{}}
Hybrid 2: LSTM  \\+ Dependency-Based Rules
\end{tabular} & 0.76 & 0.95 & 0.84 & 81.06 \\ \hline
\end{tabular}
\end{table}

To examine the effectiveness of our proposed approach with translation based approaches, we used the Google Translation API to translate our corpus into English.
The translated English corpus was evaluated using state-of-the-art sentiment analysis classifiers in English ~\cite{kim2014convolutional}.
In addition, we used n-gram based SVM and Logistic regression models to establish a baseline.

Comparative simulation results for product and hotel reviews, presented in Table \ref{tab:five-three} and Table \ref{tab:five-four} respectively, show that our proposed hybrid approach achieved better accuracy compared to CNN, LSTM, SVM and Logistic Regression classifiers.
In addition, for the product reviews dataset, the hybrid CNN approach outperforms hybrid LSTM model.
However, for the hotel reviews corpus, the hybrid LSTM approach outperforms hybrid CNN model.
It is to be noted that, the dependency-based rules are unable to classify 5\% (2\% positive and 3\% negative) of the product reviews test set and 16\% (8\% positive and 8\% negative) of the hotel reviews test set. 

Furthermore, we used the most widely used lexicon called SentiFarsNet~\cite{dehkharghani2012adaptation} to compare our dependency-based rules approach with different lexicons.
Experimental results show that the SentiFarsNet produced lower accuracy compared to PerSent lexicon. This can be attributed to the relatively smaller size of SentiFarsNet, that consists of 2500 words with polarity, whereas the PerSent lexicon comprises 3500 words.

Finally, dependency-based rules achieved better performance as compared to DNN classifiers on product reviews while DNN models outperformed dependency-based rules on product reviews. This is on account of the PerSent lexicon consisting of more words related to product reviews, compared to hotel reviews. For example \RL{gw^sy bhtr} ``Better phone''.  However, in hotel reviews, the deep learning classifiers achieved better results as compared to dependency-based rules.

\begin{table}[!t]
\centering
\caption{Summary of the results for Hotel Reviews}
\label{tab:five-four}
\setlength{\tabcolsep}{1pt}
\begin{tabular}{|l|c|c|c|c|}
\hline
\textbf{Classifier} & \textbf{Precision} & \textbf{Recall} & \textbf{F-measure} & \textbf{Accuracy} \\ \hline
Kim et al.\cite{kim2014convolutional} & 0.62 & 0.65 & 0.62 & 67.25 \\ \hline
Dehkharghani et al.\cite{dehkharghani2012adaptation} & 0.68 & 0.93 & 0.78 & 71.97 \\ \hline
fastText Classifier\cite{joulin2017bag} & 0.79 & 0.79 & 0.79 & 80.34 \\ \hline
SVM & 0.70 & 0.70 & 0.70 & 70.72 \\ \hline
Logistic Regression & 0.70 & 0.70 & 0.70 & 70.75 \\ \hline
Dependency-Based Rules & 0.80 & 0.79 & 0.79 & 79.25 \\ \hline
CNN & 0.69 & 0.92 & 0.79 & 82.33 \\ \hline
LSTM & 0.77 & 0.90 & 0.83 & 85.03 \\ \hline
\begin{tabular}[c]{@{}l@{}}
Hybrid 1: CNN \\ + Dependency-Based Rules
\end{tabular} & 0.87 & 0.91 & 0.88 & 85.91 \\ \hline
\begin{tabular}[c]{@{}l@{}}
Hybrid 2: LSTM \\ + Dependency-Based Rules
\end{tabular} & 0.87 & 0.92 & 0.89 & 86.29 \\ \hline
\end{tabular}
\end{table}

The main limitations with the dependency-based rules approach are as follows:
\begin{itemize}
    \item Dependency-based rules are unable to detect word sense disambiguation as the multi-word expressions that can help in detecting such disambiguation are absent from  the PerSent lexicon. For example, the \RL{krm} has a different meaning in Persian ``generosity'', ``worms'' and ``cream''. This types of words cannot be currently detected by the PerSent lexicon.
    \item The online reviews consist of informal words, idioms and sarcasm that cannot be detected by the PerSent lexicon. As a result the rule-based classifier cannot classify such sentences. In the future, we intend to include these words in PerSent lexicon.
    \item The dependency-based rules perform poorly on long sentences. For example, \\
    \RL{gw^sy `Aly bwd, Az^s rA.dy bwdm b`d sh mAh AstfAdeh yk hAlh qrmz Awmd rw .s.hfh b`d yk mAh brd^s swxt brdm nmAyndgy mygn seh mAh b`d bAyd .sbr kni brA brd wAq`A mtAsfm brA Aljy yk swny AkspryAl dA^stm pdr^sw dr Awrdh bwdm hy^c^s n^sd Ayn nAz nAzy Ayjwry ^sd AlAn mn mytrsm yx^cAl tlyzywn mA^syn lbAs^swyy bswzh Az Al^gy Ant.zAr ndA^stm dr .dmn gw^sy mn mdl bwd ps Alky ngyn m^skl ndArh dr xmn Ageh xwAstin Ayn gw^syw bgyryn bdwn gArAnty^s bA gArAnty hy^c frqy nmyknh pwl alky bh gArAnty ndyn AlAn yk mAh gw^sy ndArm}\\
    ``The phone was great I was really satisfied, after three months the motherboard was burned, I feel sorry for LG, I had Sony long time back, I was very happy with it, in case if you want to buy this mobile, with guarantee without guarantee it does not make any difference.''
    \item The noisy reviews data consists of numerous spelling mistakes that cannot be detected or auto corrected using a dependency-based rules approach. For example, \RL{htl xyly xw^sgz^st}, ``I had good time in hotel''. The word \RL{xw^sgz^st} ``good time'' has spelling mistakes which cannot be detected by the dependency-based rules approach.  In order to auto correct spelling mistakes, we need to incorporate a contextual spell checker in the approach.
\end{itemize}

\begin{table}[!t]
\caption{Ablation Study using Product Reviews Dataset}
\label{tab:five-nine}
\centering
\setlength{\tabcolsep}{3pt}
\begin{tabular}{|l|c|c|c|c|}
\hline
\textbf{Rules} & \textbf{Precision} & \textbf{Recall} & \textbf{F-measure} & \textbf{Accuracy} \\ \hline
Complement Clause & 0.59 & 0.67 & 0.62 & 50.32 \\ \hline
Adverbial Clause & 0.75 & 0.56 & 0.64 & 51.25 \\ \hline
Preposition & 0.53 & 0.50 & 0.51 & 51.94 \\ \hline
Adjective Clause & 0.70 & 0.58 & 0.63 & 52.47 \\ \hline
Additional rules & 0.65 & 0.57 & 0.60 & 55.62 \\ \hline
Polarity Inversion & 0.65 & 0.71 & 0.67 & 65 \\ \hline
Adversative & 0.71 & 0.68 & 0.69 & 67.54 \\ \hline
Joint Noun and Adjective & 0.68 & 0.73 & 0.70 & 68.13 \\ \hline
\end{tabular}
\end{table}

\begin{table}[!t]
\caption{Ablation Study using Hotel Reviews Dataset}
\label{tab:five-ten}
\centering
\setlength{\tabcolsep}{3pt}
\begin{tabular}{|l|c|c|c|c|}
\hline
\textbf{Rule} & \textbf{Precision} & \textbf{Recall} & \textbf{F-measure} & \textbf{Accuracy} \\ \hline
Complement Clause & 0.51 & 063 & 0.56 & 53.17 \\ \hline
Adjective Clause & 0.55 & 0.67 & 0.59 & 54 \\ \hline
Adverbial Clause & 0.53 & 0.62 & 0.56 & 54.46 \\ \hline
Additional rules & 0.61 & 0.65 & 0.62 & 56.53 \\ \hline
Preposition & 0.60 & 0.65 & 0.62 & 56.55 \\ \hline
Polarity Inversion & 0.55 & 0.56 & 0.54 & 56.56 \\ \hline
Adversative & 0.63 & 0.71 & 0.66 & 63 \\ \hline
Joint Noun and Adjective & 0.71 & 0.74 & 0.72 & 68.99 \\ \hline
\end{tabular}
\end{table}

We conduct an ablation study to better understand how each part of the system performs in isolation. Table \ref{tab:five-nine} and Table \ref{tab:five-ten} report the results of ablation studies on product reviews and hotel reviews corpora respectively. Experimental results show that the joint-noun and adjective rules achieved better accuracy in both hotel and product reviews datasets as compared to other rules. In addition, the complement clause achieved the lowest performance compared to other rules.

Table~\ref{tab:five-eleven} presents some classified sentences by the hybrid approach from product review test set.

\begin{table}[!t]
\centering
\caption {Selected examples of the Persian dependency-based rules approach}
\label{tab:five-eleven}
\begin{tabular}{|l|l|l|}
\hline
\textbf{Persian}                         & \textbf{English}                                                                         & \textbf{Polarity} \\ \hline
\begin{tabular}[c]{@{}l@{}}
\RL{wAq`A `Alyh wqty xrydm^s} \\ \RL{fhmydm .dd} \\ \RL{.drbh hst kh} \\ \RL{hy^c m^skly} \\ \RL{bA wyndwz ndArh}
\end{tabular}  & \begin{tabular}[c]{@{}l@{}}
It's really great. \\ When I bought it, \\ I realized it's a counter-hit, \\ which means there's no \\ problem with Windows
\end{tabular} & Positive     \\ \hline
\begin{tabular}[c]{@{}l@{}}
\RL{yky Az bhtryn AntxAb} \\ \RL{haa byn hArd} \\ \RL{hAy akstrnAl Ayn mdl} \\ \RL{Az Ay dytaa hst^s}
\end{tabular} & \begin{tabular}[c]{@{}l@{}}
One of the best choices \\ between the external hard \\ drives of this model is \\ the ODATA
\end{tabular}               & Positive     \\ \hline
\begin{tabular}[c]{@{}l@{}}
\RL{xyly dA.g myknh dr} \\ \RL{.hdy kh dsttwn bswzh}
\end{tabular}       & \begin{tabular}[c]{@{}l@{}}
Its getting very hot, \\ so it can damage your hand
\end{tabular}                                    & Negative     \\ \hline
\RL{hy^c ^cyz jdydy ndArh}                               & It does not have anything new & Negative     \\ \hline
\end{tabular}
\end{table}

\section{Conclusions}\label{sec:conclusion}

The quintillion bytes of data generated per day on e-commerce and social media websites, holds valuable information that can be exploited by both buyers to make informed decisions, and also by sellers to take into account past customers issues, in order to improve their products or services.
However, the data, mainly consisting of user feedback, cannot be manually read and analyzed by an individual or an organization for gauging public opinion.
Sentiment analysis offers a solution to computationally understand and classify subjective information from user generated feedback.
However, current approaches to Persian sentiment analysis are based on word co-occurrence frequencies, that fail to consider the words order and hierarchical relation between words, which are known to play an important role in determining the underlying sentiment.
In this paper, we first propose a novel framework based on Persian dependency-based rules, that consider the dependency relations between keywords, the word order and, individual word polarities to address the aforementioned issues.
In addition, we propose a novel hybrid framework for Persian sentiment analysis, that integrates dependency-based rules, and deep neural networks to address the limitation of unclassified sentences, associated with dependency-based rules.
A comparative evaluation using benchmark product and hotel reviews corpora demonstrates significant performance improvement of our proposed hybrid framework over state-of-the-art approaches based on SVM, Logistic Regression and advanced DNN models, including CNN and LSTM. Ongoing work includes the exploration of dependency-based rules to extract multilingual concepts from a mixture of Persian and English sentences, to detect subjectivity and sentiment in the sentences.
It is worth mentioning that, our proposed model has only been evaluated on reviews corpora from two domains.
In the future, we intend to address the issue of unclassified sentences by extending our lexicon, and further investigate the generalization capability of our hybrid framework using other more challenging corpora from a range of different applications including emotion sensitive companion.
Additionally, we plan to exploit Graph Neural Networks to automatically learn dependency-based rules from multilingual corpus.

\end{document}